%% file: root.tex
\documentclass[letterpaper, 10 pt, conference]{ieeeconf}

\IEEEoverridecommandlockouts
\usepackage{graphicx}
\usepackage[ruled,vlined]{algorithm2e}
\SetKwInput{KwInput}{Input}
\SetKwInput{KwOutput}{Output}
\usepackage{amsmath}
\usepackage{amssymb}
\usepackage{verbatim} 
\usepackage{hyperref}
\usepackage{flushend}

\usepackage{graphicx}
\usepackage{xcolor}

\def\Q{Q}
\def\Qfree{\Q_{\text{free}}}
\def\X{X}
\def\Xfree{\X_{\text{free}}}
\def\q{q}

\def\qstart{\q_{\text{start}}}
\def\qgoal{\q_{\text{goal}}}

\title{\Huge
Efficient Sampling of Transition Constraints for Motion Planning under Sliding Contacts}

\author{Marie-Therese Khoury$^{1}$ \and Andreas Orthey$^{2}$ \and Marc Toussaint$^{2,3}$
\thanks{$^{1}$University of Stuttgart, Stuttgart, Germany
{\tt\small marietheresekhoury@googlemail.com}}
\thanks{$^{2}$Max Planck Institute for Intelligent Systems, Stuttgart, Germany}
\thanks{$^{3}$Technical University of Berlin, Berlin, Germany}
}

\begin{document}

\maketitle
\thispagestyle{empty}
\pagestyle{empty}

\input{sections/01abstract}

\input{sections/02intro}

\input{sections/03relatedWork}

\input{sections/04background}

\input{sections/05transitions}

\input{sections/06evaluation}

\input{sections/07conclusion}



\bibliographystyle{IEEEtranS}
{\small
\bibliography{IEEEabrv, bibliography}
}
\end{document}

%% file: sections/01abstract.tex
\begin{abstract}

	Contact-based motion planning for manipulation, object exploration or balancing often requires finding sequences of fixed and sliding contacts and planning the transition from one contact in the environment to another. 
	However, most existing algorithms concentrate on the control and learning aspect of sliding contacts, but do not embed the problem into a principled framework to provide guarantees on completeness or optimality.
	To address this problem, we propose a method to extend constraint-based planning
	using contact transitions for sliding contacts.
	Such transitions are elementary operations required for whole contact sequences. 
	To model sliding contacts, we define a sliding contact constraint that permits the robot to slide on the surface of a mesh-based object.
	To exploit transitions between sliding contacts, we develop a contact transition sampler, which uses three constraint modes: contact with a start surface, no contact and contact with a goal surface.
	We sample these transition modes uniformly which makes them usable with sampling-based planning algorithms.
	Our method is evaluated by testing it on manipulator arms of two, three and seven internal degrees of freedom with different objects and various sampling-based planning algorithms. This demonstrates that sliding contact constraints could be used as an elementary method for planning long-horizon contact sequences for high-dimensional robotic systems.
	
\end{abstract}

%% file: sections/02intro.tex
\section{Introduction}

Robots that act in the real world often require the use of contacts.
Contacts are important to locomote by walking \cite{Escande2013, Orthey2013, Bouyarmane2018} or climbing \cite{B06}, to manipulate the environment \cite{Toussaint2020} or to grasp objects \cite{CHR19}. Algorithms that enable robots to move using contacts are studied in the field of contact-based motion planning \cite{Escande2013, TPM18, Kingston2019}.
The objective of contact-based motion planning is to complete a manipulation or locomotion task by reaching a goal configuration while incorporating contact constraints where chosen points on the robot have to be in contact with chosen surfaces in the environment. 

One approach to deal with contact constraints is projection-based constraint planning \cite{Berenson2011, Kingston2019}. In projection-based constraint planning, we implicitly define contact constraints, which we can sample using dedicated projection methods \cite{Berenson2011}. Such a framework has been used to plan contact motions for the Robonaut 2 robot \cite{Kingston2019}. While projection-based planning works well for fixed contact planning \cite{Kingston2019}, there does not yet exist an extension to incorporate sliding constraints or transitions between sliding contacts. Sliding contacts however, are a fundamental requirement if we want to explore the shape of an object \cite{DET17}, adjust an object grasp \cite{Shi2017}, keep the robot in balance \cite{Samadi2020}, minimize state uncertainty \cite{Pall2018}, plan more efficiently \cite{Roussel2019} or animate computer characters \cite{Hu2019}. 

To overcome those limitations of constraint planning, we develop an extension of constraint planning for sliding contacts. This extension consists of the following items:

\begin{enumerate}
    \item Definition of sliding contact constraint and transition constraint by extending the constraint-based planning framework \cite{Kingston2019}.
    \item Sampling method for the transition constraint, which uses constraint modes for efficient sampling.
    \item Evaluations with various sampling-based planning algorithms for one-step contact cycles using manipulator arms with two, three and seven internal degrees of freedom (dof).
\end{enumerate}

\begin{figure}[t]
\centering
	\includegraphics[width=\linewidth]{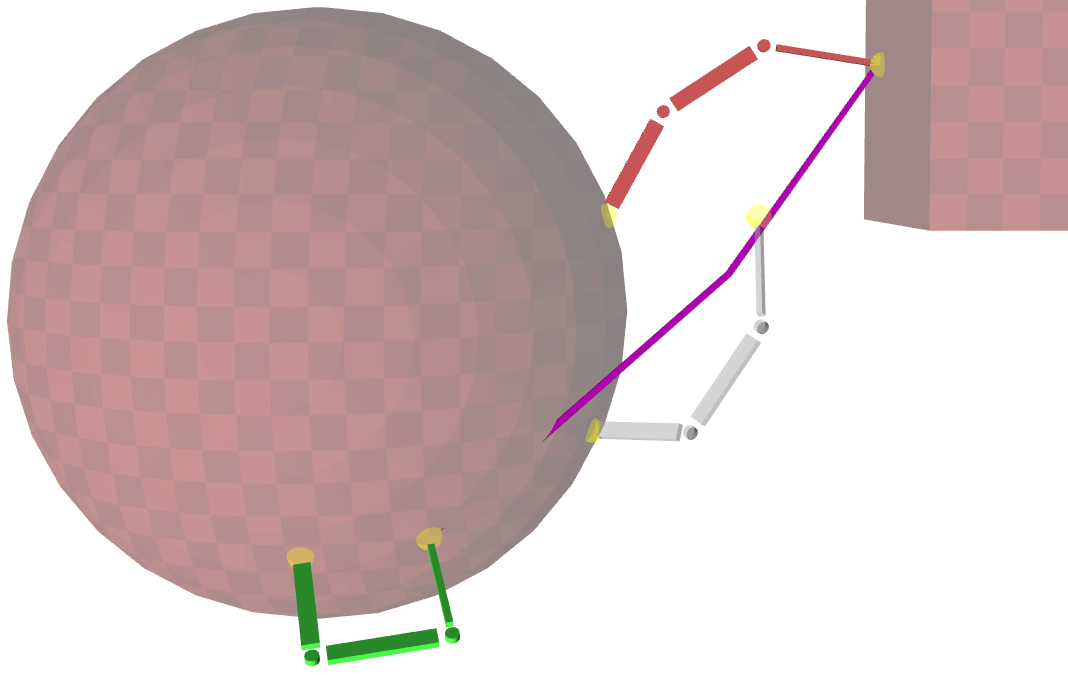}
	\caption{We present a method to plan constraints for sliding contact and transition motions on arbitrary mesh surfaces, where predefined contact points of a robot (yellow) are required to stay in constant contact with a surface or to transition.}
	\label{fig:example}
\end{figure}

Note that we concentrate in this paper exclusively on one-step sliding contact cycles, because solving such cycles robustly is of fundamental importance for planning long-horizon contact sequences \cite{Toussaint2020}. Furthermore, while our approach makes simplifying assumptions on the control of sliding contacts, our motivation is to eventually have a concise framework in which we can guarantee completeness and optimality. As an example, Figure \ref{fig:example} shows a one-step sliding contact and transition cycle for a robot with three internal dof and two contact points, where we exhaustively sample the transition constraints of sliding contacts to find a solution path.

%% file: sections/03relatedWork.tex
\section{Related Work}

Planning contact sequences through an environment can be addressed using constraint-based motion planning \cite{Berenson2011, Kingston2019}. This approach is based on building probabilistic road maps or trees by randomly sampling configurations, which are projected onto the contact constraints to make them feasible \cite{Berenson2011}. The transition between contact constraints is often captured using a constraint graph \cite{Mirabel2017}, through which we can identify sequences of contacts to reach a goal \cite{Escande2013}. Such sequences are often used to plan more complicated contact or manipulation tasks \cite{Garrett2020, Toussaint2020}. Because contact constraints can often not be described analytically, we have to either learn them \cite{Fernandez2020} or use methods to project
onto them \cite{Berenson2011}. To plan longer contact sequences, it often becomes necessary to exploit the structure of the state space. Different approaches exist, for example by factoring the state space for more efficient sampling \cite{Vega2020}, by computing guide paths from a simplified robot geometry \cite{TPM18} or by precomputing convex regions in workspace which we can exploit using mixed-integer programming \cite{Deits2014}. Our approach is complementary, in that we add sliding transition constraints to the constraint-based framework \cite{Kingston2019}, which makes it possible to plan sliding transition constraints using sampling-based planner \cite{Lav06} or asymptotically optimal planner \cite{Karaman2011}. 

While constraint-based planning is often used to plan non-sliding contacts, there exists substantial research on the control aspect of sliding motions \cite{Trinkle1990, Cruciani2020}. To execute and control sliding motions, we could use physical models which make use of friction cone models \cite{Samadi2020} to directly optimize sliding motions to push objects along planar surfaces \cite{Toussaint2020}. An alternative approach would be to learn the parameters of a controller for specific object categories \cite{Li2019}. Recent results using learned sliding controllers can be used for dexterous in-hand manipulation, where objects are either rotated in-hand without making contact with the environment \cite{Kumar2016, Andrychowicz2020} or features in the environment are used as helper tools \cite{Cruciani2020}. Our approach is complementary, in that we make simplifying assumptions on the control of sliding contacts (deterministic, no uncertainty) and concentrate on the computational challenges of planning sliding contacts \cite{Lee2015}, which could be used to provide guarantees on completeness and optimality.

%% file: sections/04background.tex
\section{Background}
This section provides some background knowledge.
Large parts of this section are based on Part II of \cite{Lav06} and \cite{KMK18}.

A \textbf{configuration} $q$ defines the independent variables of a given robot needed to uniquely specify the robot's position relative to a reference frame.
Considering a manipulator arm with $m$ number of rigid links connected by $n$ joints that is fixed at one end, one possible configuration would define $n$ internal variables plus external variables to describe a root frame fixed to the robot.
The \textbf{configuration space} $\Q$ of a robot is the set of all such possible configurations $q$.
To avoid collisions with any obstacles or the robot itself, typically a free space $\Qfree \subseteq \Q$ is defined that only contains collision-free configurations. 

A basic motion planning problem is defined as finding a continuous path in $\Qfree$ that connects from a given start configuration $\qstart  \in \Qfree$ to a goal configuration $\qgoal \in \Qfree$.
Explicitly computing $\Qfree$ is a complex problem and the complexity grows with an increase of a robot's degrees of freedom \cite{Canny1988}.
We use the concept of sampling-based motion planning to avoid this by working with an implicit representation of $\Qfree$ probed through sampling strategies.

\textbf{Sampling-based motion planning} \cite{Lav06} relies on sampling collision-free configurations $q$ in $\Qfree$ which are connected to a tree or a graph. 
With a longer planning time and increasing numbers of samples, more space can be mapped out.
If the problem is solvable, the probability of finding a path then converges to one if time goes to infinity.
The algorithms provide an efficient solution for finding feasible paths for high-dimensional problems.

\textbf{Constraint-based planning} can be used whenever a robot performs tasks that limit its possible motions to subspaces of the configuration space \cite{KMK18}.
When we use a contact-based method with point contacts, sliding contacts or contact transitions, we put such limits on a robot's configuration space $Q$.
In order to express these limits, we formulate a contact as a single task constraint and a transition as a set of successive task constraints.
We extend the motion planning objective of finding collision-free configurations to simultaneously satisfy given constraints.
These constraints are specified by a constraint function $$f(q): Q \rightarrow \mathbb{R}^{n}$$ that is satisfied when the real-value vector  $f(q) = \textbf{0}$ for a given configuration $q$.
This function is then used to construct an implicitly constrained configuration space. $$X = \{q \in Q \mid f(q) = \textbf{0}\}$$
It contains all configurations that satisfy the defined constraint.
We can now define a configuration space that includes all collision-free configurations from $\Qfree$ that also satisfy the constraints.
$$\Xfree = X \cap \Qfree$$
A basic constrained motion planning problem is thus defined as finding a continuous path in $\Xfree$ that connects from a given start configuration $\qstart  \in  \Xfree$ to a goal configuration $\qgoal \in  \Xfree$.

We assume that a robot has $k$ \textbf{contact points}, which are predefined points on the robots geometry. 
To each contact point, we associate a mapping $f_k$ which maps a joint configuration and a surface to the distance of the $k$-th contact point to the nearest point on the surface. 
For a surface $s$ in a workspace $W$, we say that the contact point fulfills a contact constraint if $f_k(q,s)=0$. 

A continuous motion for a contact point which fulfills the contact constraint over a surface is a \textbf{sliding contact}.
A \textbf{contact transition} is the process of breaking a contact on an initial surface $s_I$, moving freely and creating a contact on a goal surface $s_G$. 
In configuration space, we thereby transition from the constraint surface of all configurations $q$ in $Q$ for which $f_k(q,s_I)=0$ to being constraint-free to being in the set of configurations for which $f_k(q,s_G)=0$. 
To keep track of sliding constraints and their transitions, we make use of the concept of a constraint graph \cite{Mirabel2017, Kingston2019}, which is a graph with vertices being active constraints and edges being possible transitions between active constraints. In the next section we elaborate on such a constraint graph.

%% file: sections/05transitions.tex
\section{Sliding Contacts and Transition Constraints}

The goal of this work is realizing a new method for planning a one-step contact cycle for sliding contacts and the transition between two given surfaces. For this purpose, we develop a sliding contact constraint, a transition constraint and a sampler to efficiently exploit the transition constraint.

The sliding contact constraint keeps a contact point in constant touch with the surface of an object, while the transition constraint encompasses a contact break from a given start surface, the contact-free motion toward a goal surface and a contact creation at the goal.
The steps of a transition motion are divided into three separate transition modes.
We combine the concept of our contact constraint with a mode sampler that samples these three modes uniformly.

For a robot with $k$ contact points we predefine each to be either a constant sliding contact point or one that performs a transition and we specify $k$ constraints accordingly.
Only these $k$ designated contact points can be in contact, the rest of the body is planned to avoid any collision.

\subsection{Explanatory Example}

\begin{figure}
    \centering
    \includegraphics[width=\linewidth]{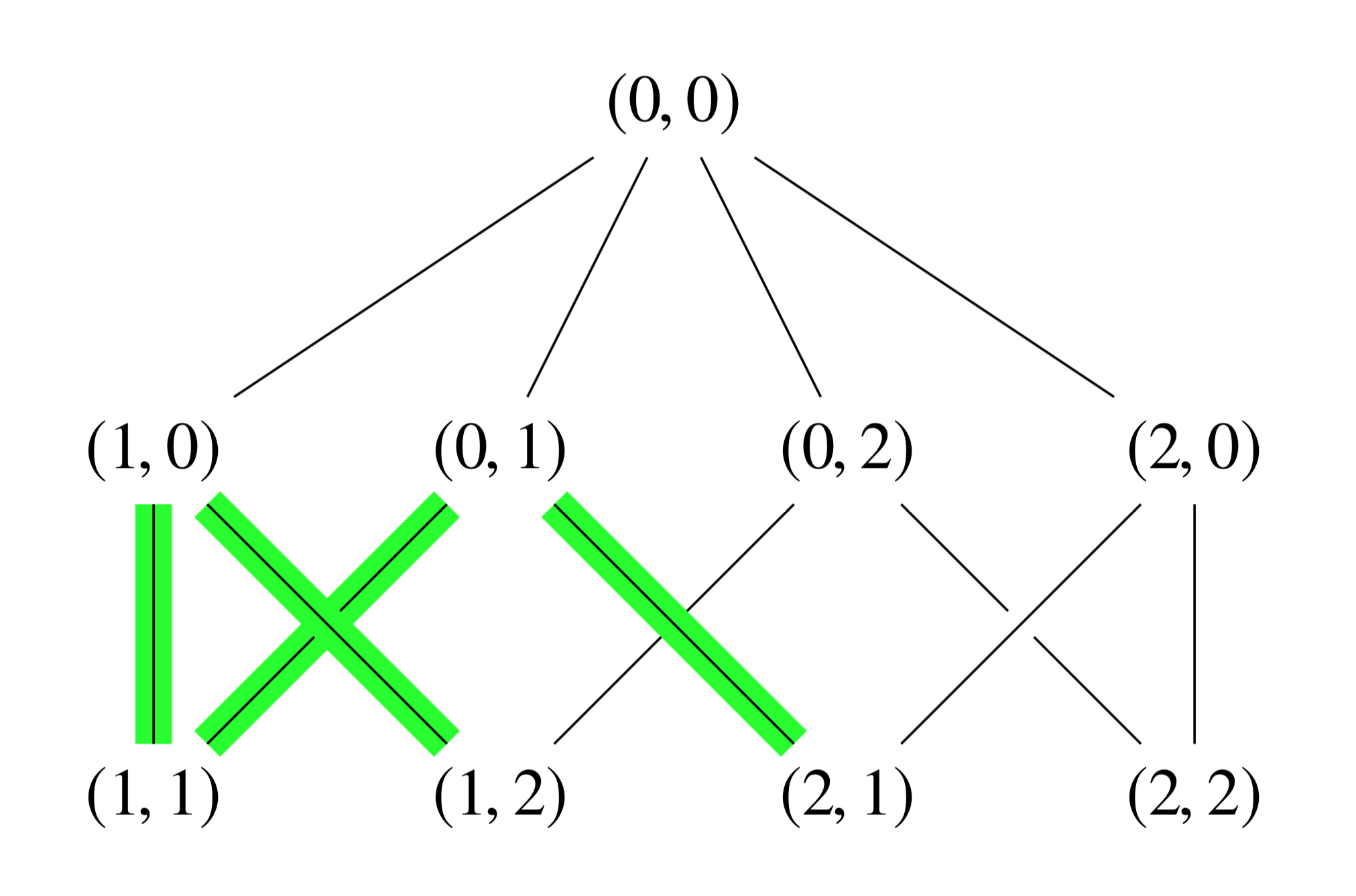}
    \caption{Constraint Graph. This graph depicts the possible modes and mode changes of a robot with $k = 2$ contact joints and two contact surfaces. Relevant for this work are the parts highlighted in green.}
    \label{fig:constraintGraph}
\end{figure}

\begin{figure}
	\centering
	\includegraphics[width=0.48\linewidth]{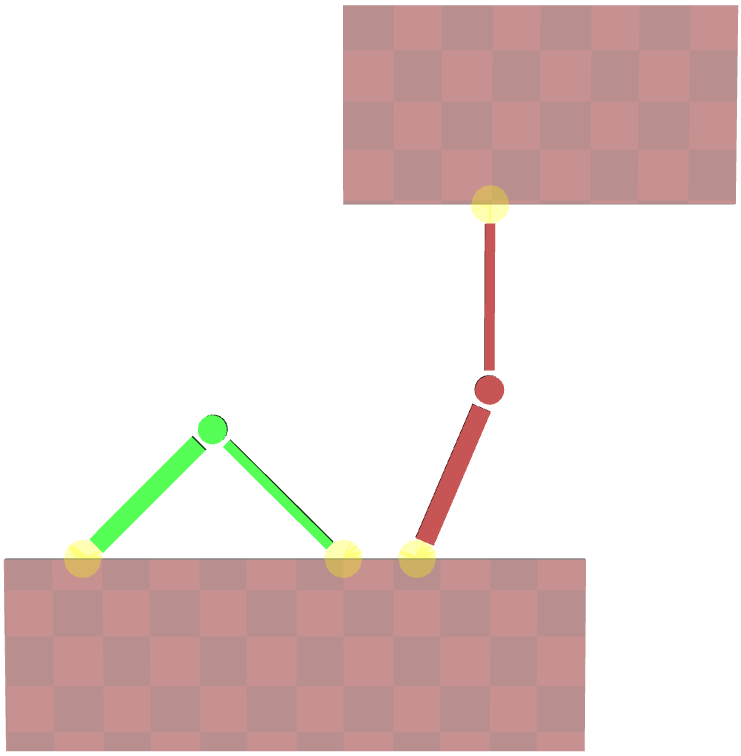}
	\includegraphics[width=0.48\linewidth]{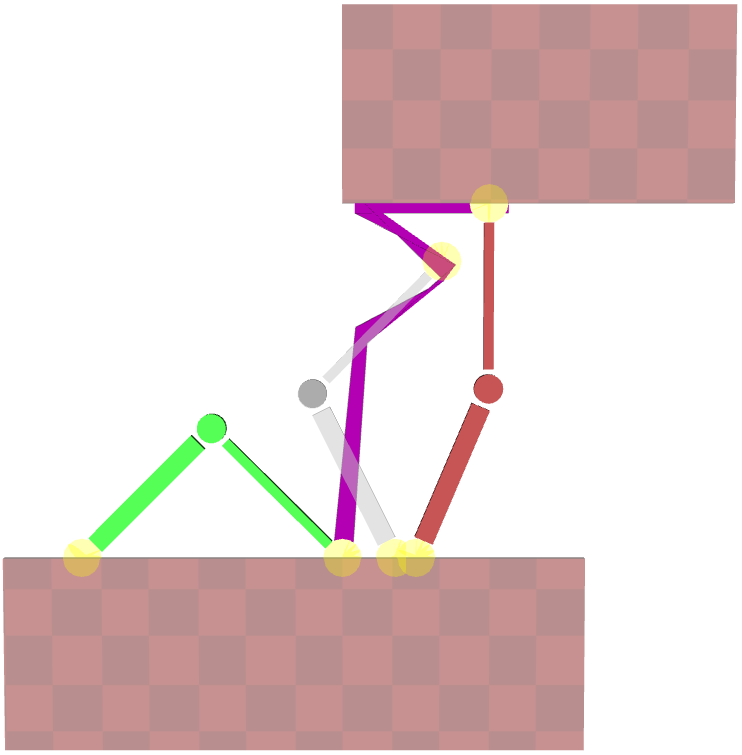}
	\caption[Two dof manipulator rectangle scenario]{Left: Two dof rectangle scenario. $q_{start}$ is depicted in green and $q_{goal}$ in red. Right: Path with intermediate step.}
	\label{fig:2dof}
\end{figure}

In Figure~\ref{fig:constraintGraph} we show a constraint graph example for a robot with $k = 2$ contact points and two contact surfaces. This graph corresponds to the situation of the robot in Figure~\ref{fig:2dof}, which is initially in contact with surface one (bottom box) and has to transition one contact point to surface two (top box) while the other contact point can slide along surface one.
The constraint graph depicts the three different modes a robot's contact point can be in and which modes can be accessed afterwards.
The $k$-th entry of a tuple corresponds to the assignment of the $k$-th contact point to the according surface or to be free. In our example, each contact point can thus be in one of three modes.

\begin{itemize}
	\item \textbf{Mode 0} contact point is unconstrained.
	\item \textbf{Mode 1} contact point is in contact with surface one.
	\item \textbf{Mode 2} contact point is in contact with surface two.
\end{itemize}

State (0,0) describes a free floating robot without any contacts.
From there either contact point can make contact with object surface one or two. 
This mode change is depicted by the edges between the different states.
State (1,1) corresponds to a robot with both contacts on the same initial surface $1$ (see the green stance in Figure \ref{fig:2dof}) and state (2,2) corresponds to a double contact with surface $2$.
A step cycle describes the process of a joint breaking contact with one surface and creating a new contact on another surface. 
At least one of the two joints is in contact with one of the surfaces during the whole process.
A step cycle is completed by changing modes along the depicted edges starting for example at state (1,1) and ending in state (2,1).
From there, the cycle can be repeated to form a step sequence of contacts and transitions.

Highlighted in green are the two particular one-step cycle examples. Starting from state (1,1), we consider the task of moving along the edges over (1,0) or (0,1) to state (1,2) or state (2,1), respectively. 
This corresponds to breaking the first respectively second contact point from the first surface, freely moving it and finally making contact with the second surface while the second respectively first contact point stays in sliding contact with the first surface.


\subsection{Sliding Constraint\label{chap:contactconstraint}}

First we present the sliding contact constraint that is put on the $k$-th contact point of the robot specified to be in sliding contact.
We compute the distance between this contact point and the closest point on a given surface $S$ at the current configuration $q$ and assign this value to the constraint function $f_k(q,S)$.

Algorithm \ref{alg:contact} shows the pseudo-code for computing sliding constraints.
The inputs are the robot $r$, the contact point on the robot $p$ and the given contact surface $S$.
First the contact point's position in world coordinates $p_w$ is determined.
The distance $d$ is first declared infinite.
We iterate over partitions of the whole $S$ (meaning the triangles of a given surface mesh).
During each iteration we save the coordinate on one such partition $s$ that is closest to $p_w$ into $c$ and the distance between these two coordinates into the variable $d_s$.
We then compare the two distance variables and if $d_s$ is smaller than $d$, we assign the value of $d_s$ to $d$.
After this loop, we have determined the coordinate of the surface partition closest to the current position of the robot contact joint and return this smallest distance $d$.

\begin{algorithm}
 \KwInput{Robot $r$, Contact point $p$, Surface $S$}
 \KwOutput{Distance d}
 $p_w \leftarrow getWorldPosition(r, p)$\;
 $d = +\infty$\;
 \For{$s\text{ in }S$}{
  $c \leftarrow closestPosition(s, p_w)$\;
  $d_s \leftarrow getDistance(c, p_w)$\;
  \uIf{$d_s < d$}{
    $d \leftarrow d_s$\;
   }
  \textbf{return} $d$\;
 }
 \caption{Sliding Constraint}
 \label{alg:contact}
\end{algorithm}

\subsection{Transition Constraint}
This section presents the implementation of our transition constraint.
It combines the sliding contact constraint with a constraint-free state into three transition modes.

Algorithm \ref{alg:transition} shows the pseudo-code for the transition constraints.
It requires input on the robot $r$, the contact point on the robot $p$, the transition mode $mode$, the desired contact start surface $S_I$ and the goal surface $S_G$.
First the contact point's position in world coordinates $p_w$ is determined.
The distance $d$ is first declared infinite.
The transition $mode$ of the current iteration is set by our sampling method explained in the next section.
We enforce a different constraint depending on the mode.
In the case of $mode = 0$ the returned distance $d$ is 0.
This corresponds to a contact-free state and by definition the constraint is satisfied.
If $mode = 1$ then we compute a sliding contact constraint using surface $S_I$.
The last case of $mode = 2$ computes a sliding constraint on surface $S_G$.

\begin{algorithm}
 \KwInput{Robot $r$, Contact point $p$, Mode $mode$, Start surface $S_I$, Goal surface $S_G$}
 \KwOutput{distance d}
  $p_w \leftarrow getWorldPosition(r, p)$\;
  \uIf{$mode = 0$}{
    \textbf{return} $d = 0$
   }\uElseIf{$mode = 1$}{
    $d \leftarrow slidingConstraint(r, p, S_I)$\;
    \textbf{return} $d$\;
  }\uElseIf{$mode = 2$}{
    $d \leftarrow slidingConstraint(r, p, S_G)$\;
    \textbf{return} $d$\;
 }
 \caption{Transition Constraint}
 \label{alg:transition}
\end{algorithm}

\subsection{Sampling Method}

In order to use the transition constraint for path planning, we implement a method to sample the three different constraint modes.
The mode is set before calling the transition constraint algorithm and the transition constraint is enforced accordingly.

In Algorithm \ref{alg:sampler} we show the pseudo-code for sampling transition modes.
The input is the configuration space $\Q$ and a list of the $k$ specified constraints $C$. Each constraint could be either a sliding or a transition constraint. We first sample a configuration $q$ uniformly in $\Q$. 
We then iterate over the constraints and check each entry if it is a transition constraint.
For each transition constraint we then assign a uniformly sampled integer from $[0, 1, 2]$ to the transition $mode$. Once the modes are set, we call the projection method which minimizes the constraint distance. Details on the project method can be found in \cite{Kingston2019}. To verify that we successfully projected, we then compute the distance and return the projected configuration. 

\begin{algorithm}
 \KwInput{Configuration Space \Q, Constraints C}
 \KwOutput{Configuration q, Distance d}
 \SetAlgoLined
 $q \leftarrow sampleUniform(\Q)$\;
 \For{$c\text{ in }C$}{
  \uIf{$c$  \textbf{is}  $TransitionConstraint$}{
   $c.mode \leftarrow uniformSampledInt(0, 2)$\;
   }
 }
 $q \leftarrow C.project(q)$\;
 $d \leftarrow C.distance(q)$\;
 \textbf{return} $q, d$\;
 \caption{Transition Mode Sampler}
 \label{alg:sampler}
\end{algorithm}

%% file: sections/06evaluation.tex
\section{Evaluation and Results}

We test our transition constraints with manipulator arms of two, three and seven internal dof and different obstacle scenarios in two-dimensional (2D) and three-dimensional (3D) space.
The scenarios are run ten times with each planning algorithm and with a specified computation time. 
We compare five sampling-based planning algorithms, which performed well in initial experiments. In detail, we use Rapidly Exploring Random Tree (RRT) \cite{Kuffner2000}, Sparse Stable RRT (SST) \cite{Li2016}, Search Tree with Resolution Independent Density Estimation (STRIDE) \cite{Gipson2013}, Probabilistic Roadmap Method (PRM) \cite{kavraki_1996} and Sparse Roadmap Spanners (SPARS) \cite{dobson_2014}. We evaluate the performance of our method by comparing the average computation times of the different algorithms.

\subsection{Programming Setup}

The sliding contact and transition constraints and sampler are implemented as extension of the constraint-based planning framework \cite{Kingston2019} of the Open Motion Planning Library \cite{OMPL, OMPLBenchmark}. To simulate and visualize the output, we use Kris' Locomotion and Manipulation Planning Toolbox \cite{Klampt}. The code is written in the programming language C++ and is freely available\footnote{\url{https://github.com/mtkhoury/MotionPlanningExplorerGUI/tree/contact_feature}}.

\subsection{Two dof Manipulator Arm}

Our first scenario is a 2D scenario with two rectangular objects and a manipulator arm with two rigid links, two joints and two contact points (yellow) to make contacts with (Figure \ref{fig:2dof}).
The robot starts in full contact with the lower rectangle and is required to slide along the surface with the first contact point (left), while making a sliding transition with the second contact point (right). A computed path is depicted in purple. In its goal position the robot keeps one contact with the initial surface and makes one contact with the upper rectangle.

\begin{figure}[hpbt]
	\centering
	\includegraphics[width=\linewidth]{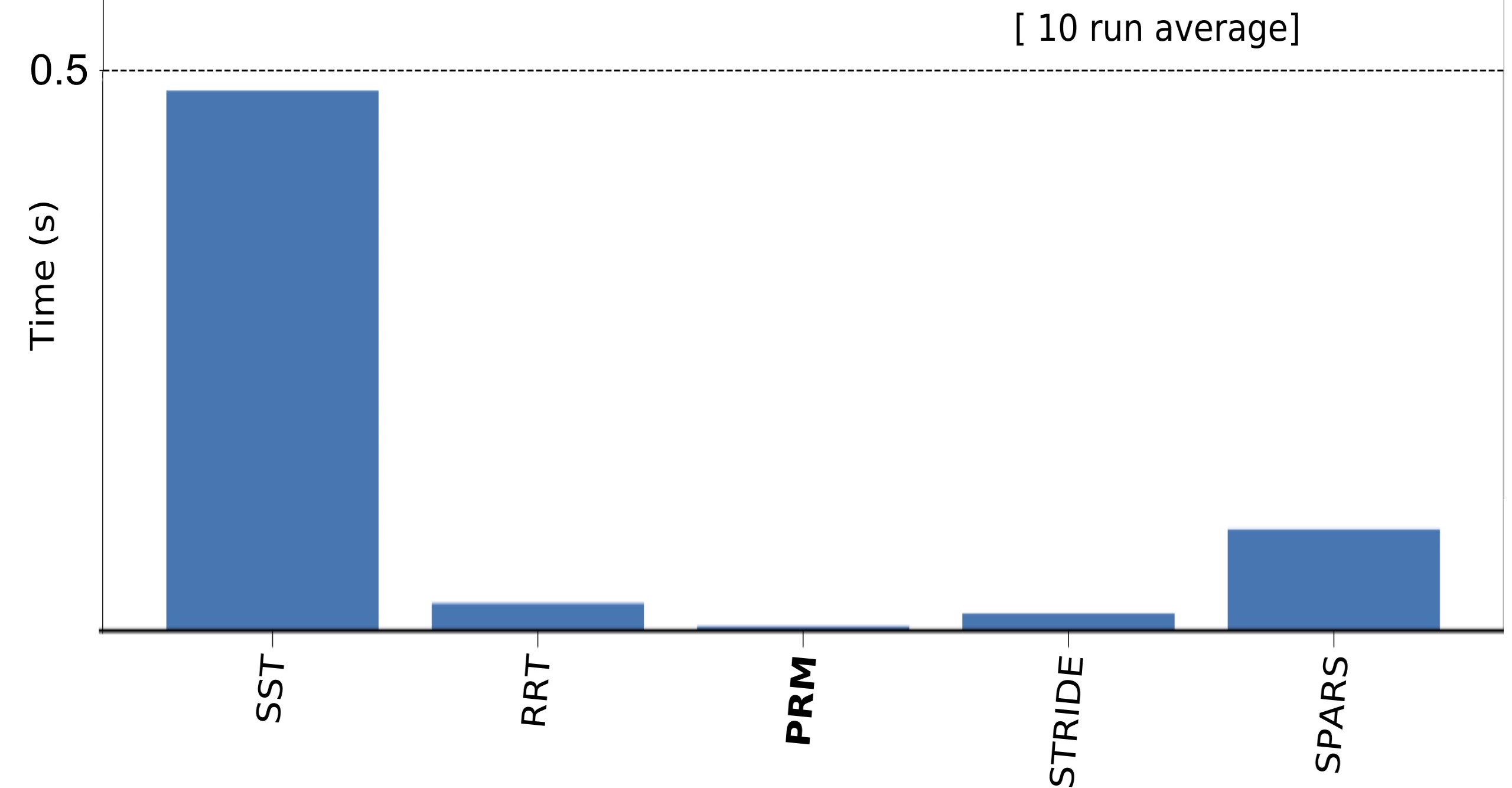}
	\caption[Computation Time Graph of Figure \ref{fig:2dof}]{Computation Time. Two dof Scenario. Maximum planning time of 0.5 seconds.}
	\label{fig:time2dof}
\end{figure}

Figure \ref{fig:time2dof} shows the average computation time of ten runs of this scenario with a specified maximum planning time of 0.5 seconds. 
We can see that all planning algorithms successfully find a path in the allocated time. 

\subsection{Seven dof Manipulator Arm}

\begin{figure}[hpbt]
	\centering
	\includegraphics[width=0.48\linewidth]{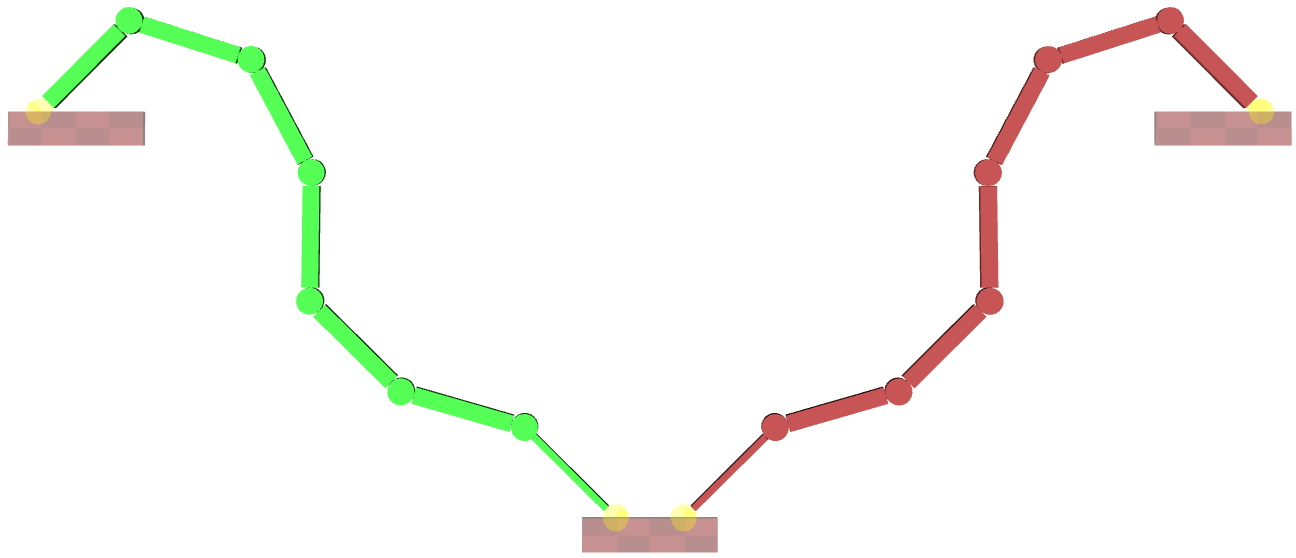}
	\includegraphics[width=0.48\linewidth]{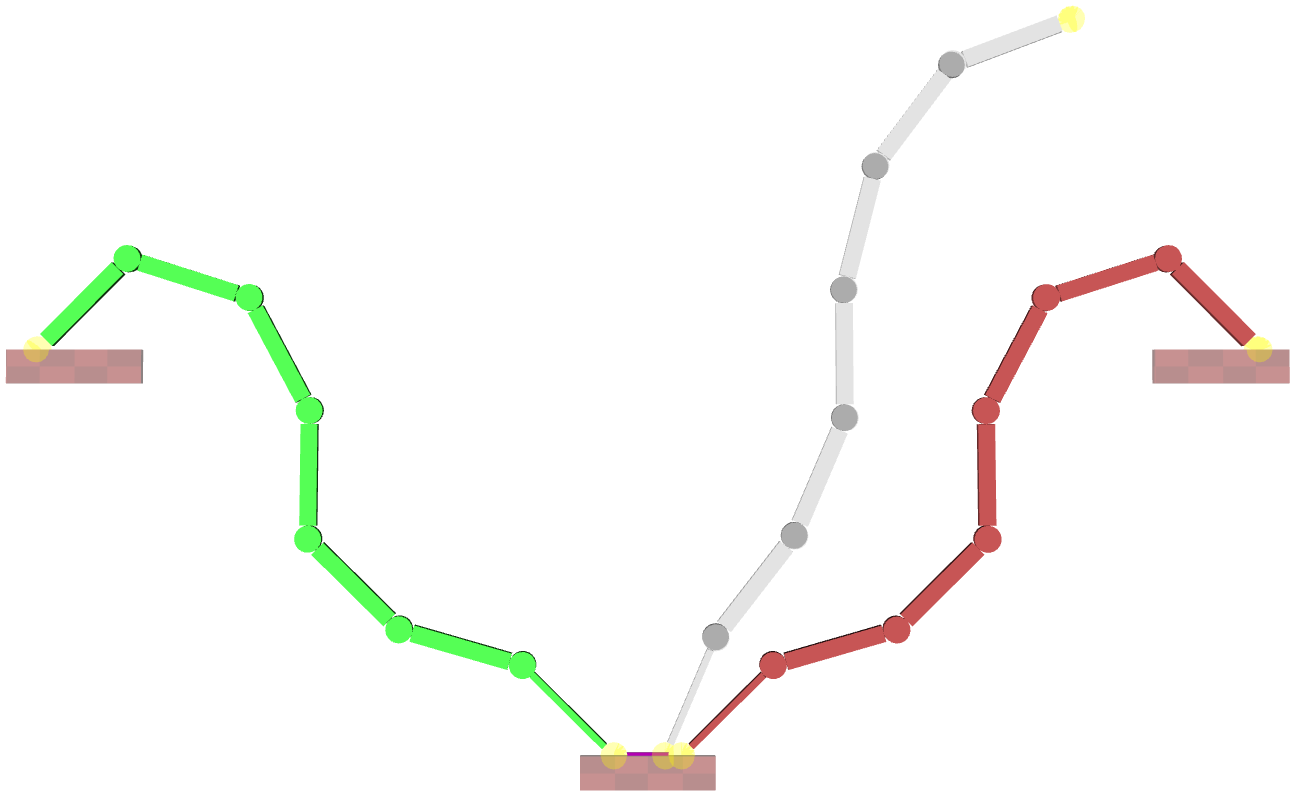}

	\caption[Seven dof manipulator rectangle scenario]{Seven dof manipulator rectangle scenario. $q_{start}$ is depicted in green and $q_{goal}$ in red.}
	\label{fig:7dofCube}
\end{figure}

The next scenario contains three rectangular objects and a manipulator arm with seven joints and two contact points (Figure \ref{fig:7dofCube}).
The robot starts with one contact point on the lower rectangle and the other contact point on the left-hand rectangle.
The task is to transition the upper end joint toward the right-hand rectangle.
To its goal position the robot has to slide the lower contact point along the lower rectangle's surface and has to make contact with the upper contact point on the upper right rectangle.

\begin{figure}[hpbt]
	\centering
	\includegraphics[width=\linewidth]{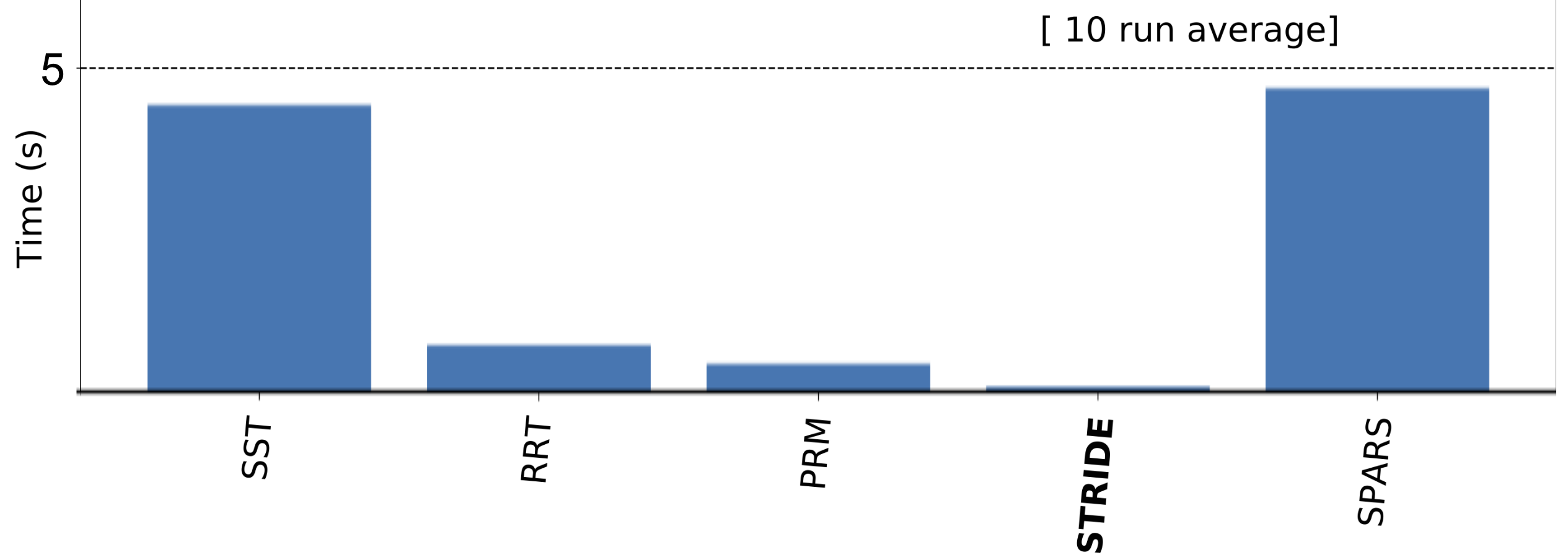}
	\caption[Computation Time Graph of Figure \ref{fig:7dofCube}]{Computation Time. Seven dof Scenario. Maximum planning time of 5 seconds.}
	\label{fig:time7dofCube}
\end{figure}

Figure \ref{fig:time7dofCube} shows the ten run average computation time graph of this scenario with a specified maximum planning time of 5 seconds. 
We can observe that all planning algorithms find a path successfully in five seconds or less (on average). Both SPARS and SST require more time in comparison to the other three planners but they still solve the problem quickly.

\subsection{Three dof Manipulator Arm on Sphere}

\begin{figure}[hpbt]
	\centering
	\includegraphics[width=0.48\linewidth]{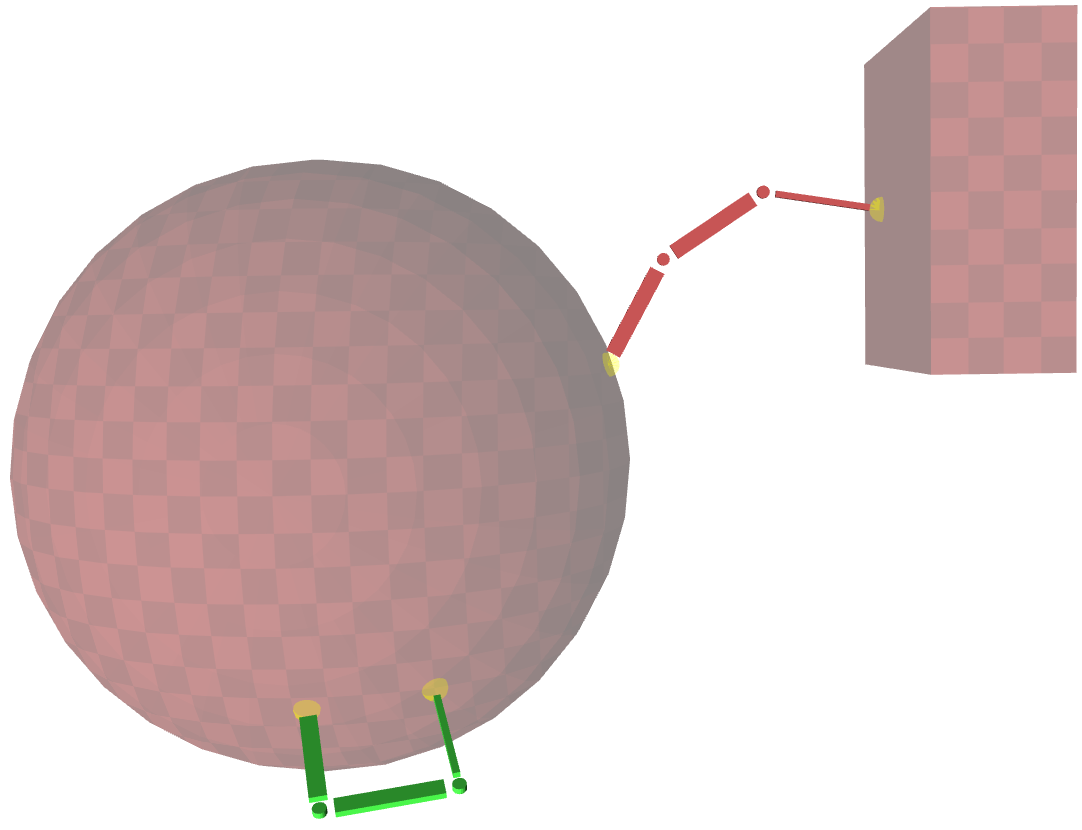}
	\includegraphics[width=0.48\linewidth]{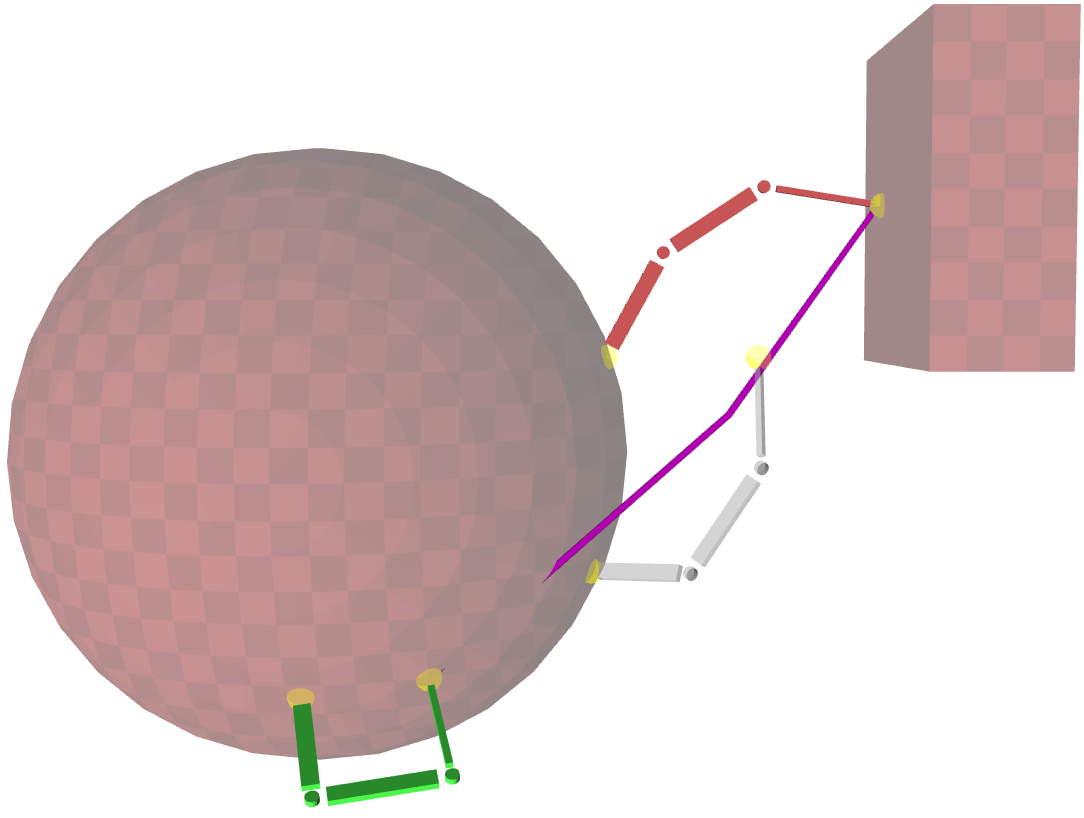}
	\caption[Three dof manipulator sphere scenario]{Left: Three dof cuboid scenario. $q_{start}$ is depicted in green and $q_{goal}$ in red. Right: Path with intermediate state.}
	\label{fig:3dofSphere}
\end{figure}

The last scenario contains a sphere, a cuboid object and a manipulator arm with three dof (Figure \ref{fig:3dofSphere}).
The robot starts in full contact with the sphere and has to transition up along the sphere to make contact with the upper right cuboid. During the transition, we restrict one contact point to remain in sliding contact with the sphere. A solution path is shown on the right.

\begin{figure}[hpbt]
	\centering
	\includegraphics[width=\linewidth]{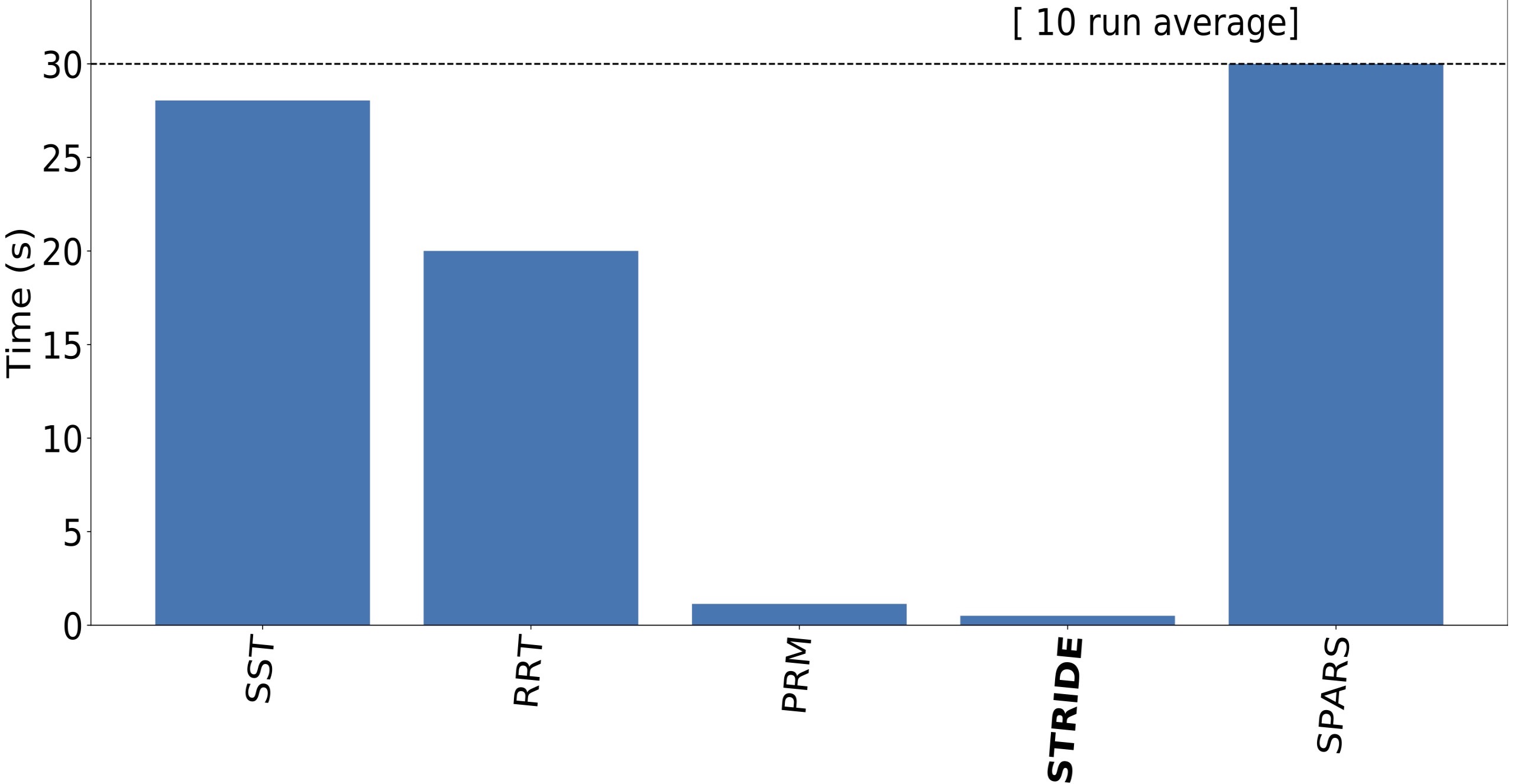}
	\caption[Computation Time Graph of Figure \ref{fig:3dofSphere}]{Computation Time. Three dof sphere scenario. Maximum planning time of 30 seconds.}
	\label{fig:time3dofSphere}
\end{figure}

Figure \ref{fig:time3dofSphere} shows the average computation time of ten runs of this scenario with a specified maximum planning time of 30 seconds. 
We can see that not all planning algorithms find a path in the maximum planning time.
The algorithms STRIDE and PRM solve the planning problem the fastest in under 5 seconds and SPARS exceeds the time limit.

\subsection{Limitations}

\begin{figure}[hpbt]
	\centering
	\begin{minipage}{0.485\linewidth}
		\centering
		\includegraphics[width=\textwidth]{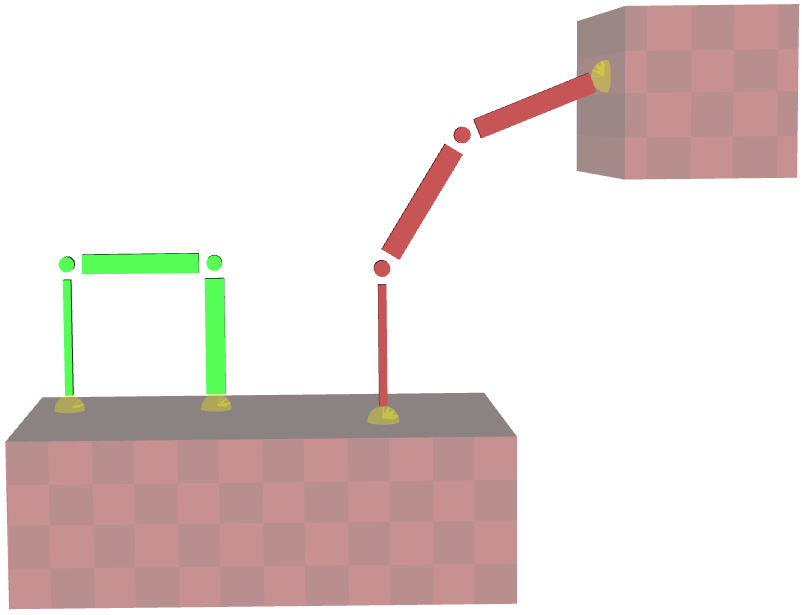}
		\caption[Three dof manipulator First Joint Sliding]{Three dof with first joint in sliding contact. $\qstart$ is depicted in green and $\qgoal$ in red.}
		\label{fig:firstFixed}
	\end{minipage}\hfill
	\begin{minipage}{0.485\linewidth}
		\centering
		\includegraphics[width=\textwidth]{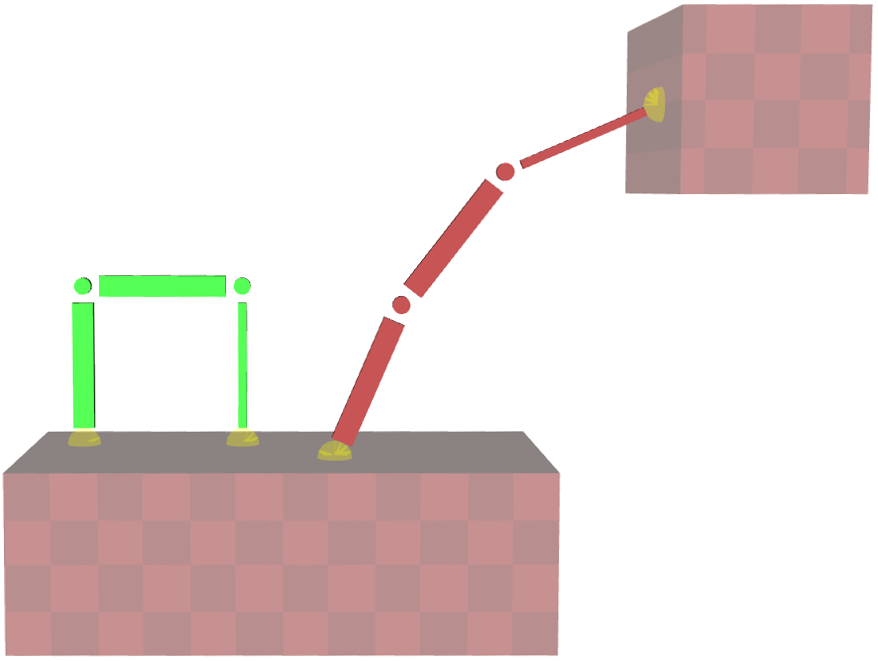}
		\caption[Three dof manipulator Last Joint Sliding]{Three dof with last joint in sliding contact. $\qstart$ is depicted in green and $\qgoal$ in red.}
		\label{fig:lastFixed}
	\end{minipage}
\end{figure}

\begin{figure}[hpbt]
	\centering
	\includegraphics[width=\linewidth]{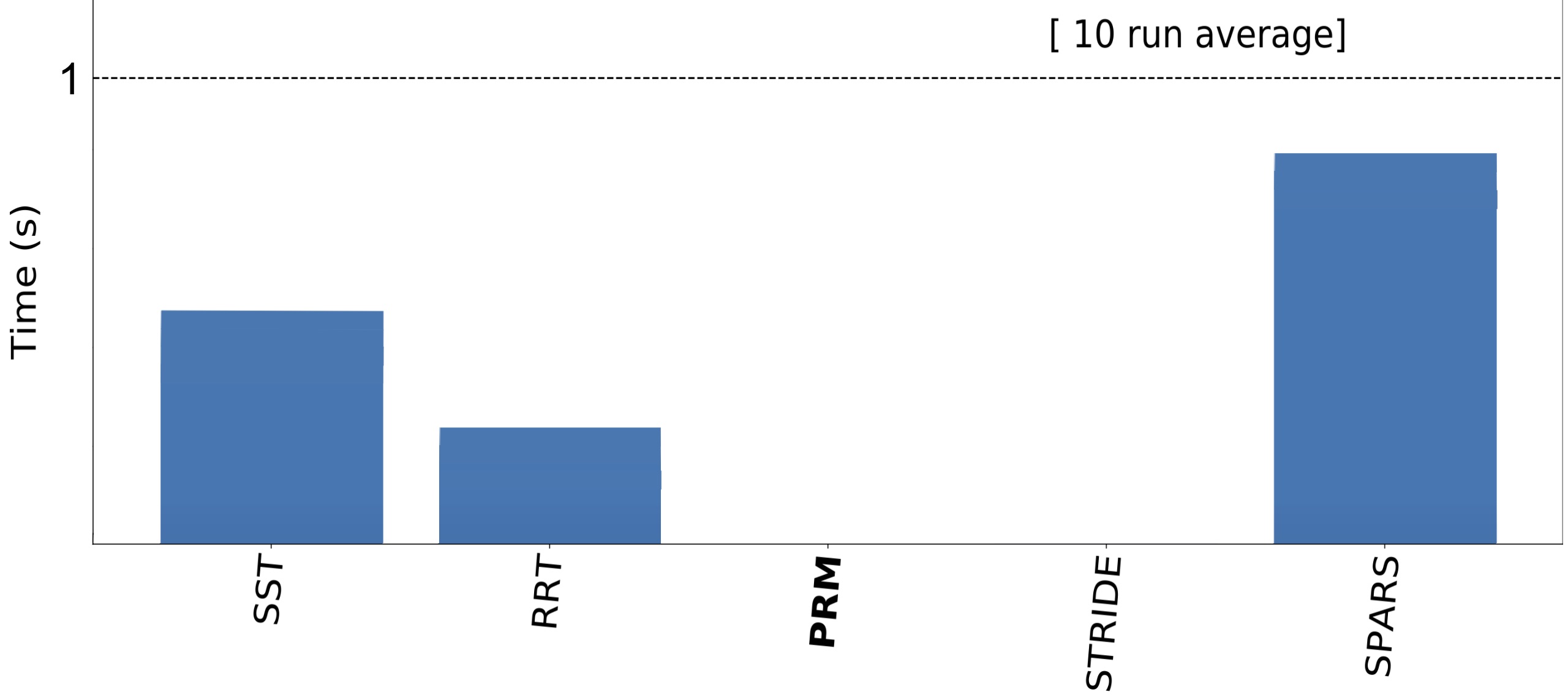}
	\caption[Computation Time Graph Figure of \ref{fig:firstFixed}]{Computation times for the first joint in sliding contact. Maximum planning time of 1 second.}
	\label{fig:timeFirstFixed}
\end{figure}

\begin{figure}[hpbt]
	\centering
	\includegraphics[width=\linewidth]{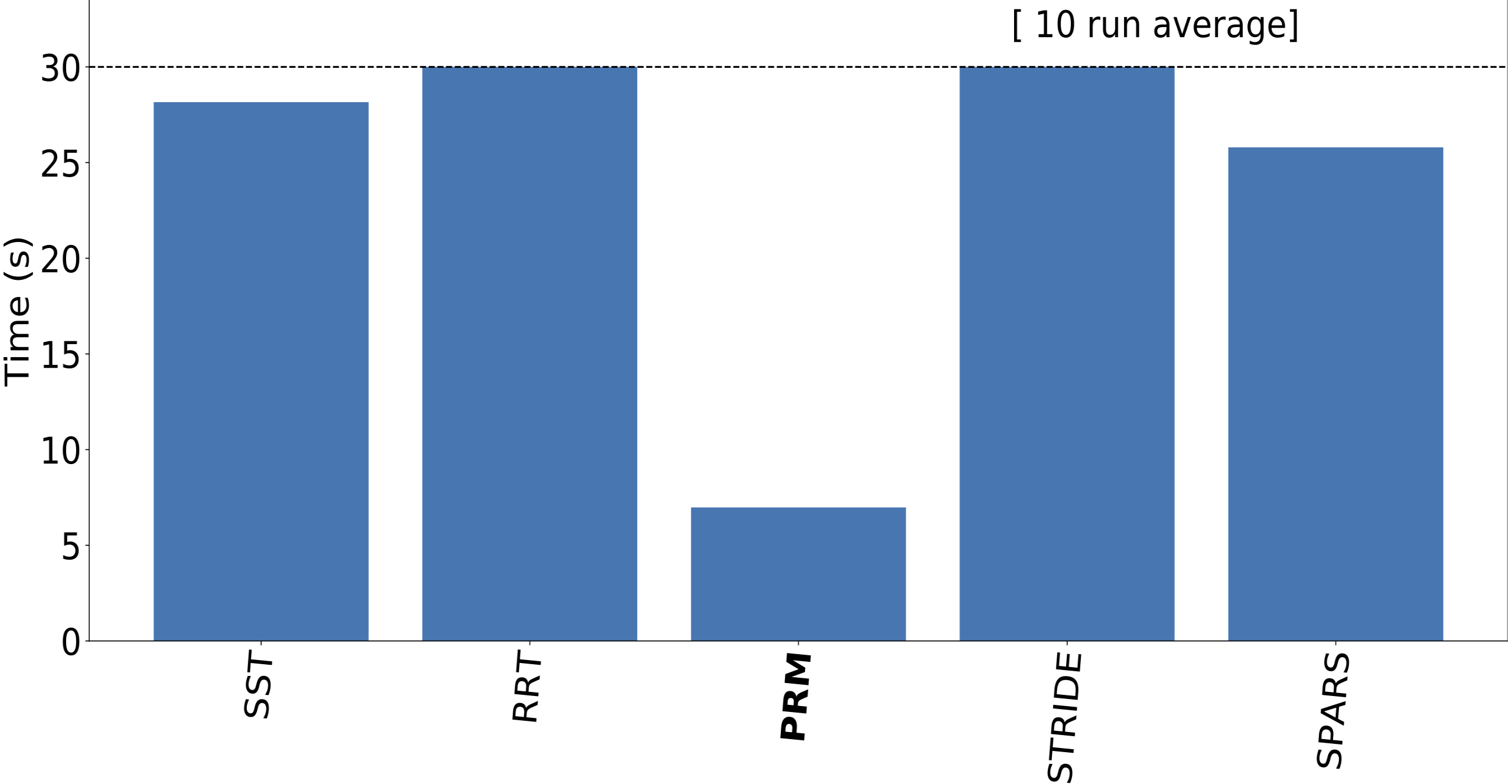}
	\caption[Computation Time Graph Figure of \ref{fig:lastFixed}]{Computation times for the last joint in sliding contact. Maximum planning time of 30 seconds.}
	\label{fig:timeLastFixed}
\end{figure}

During the setup of our testing scenarios we came across a limitation due to the different ways to set the projection.
One setup, shown in Figure \ref{fig:firstFixed}, uses the first contact point (end of thinner link) to stay in contact with the initial surface while the last contact point (end of thicker link) transitions.
The other setup is vice-versa, the first contact point transitions and the last contact point keeps contact (Figure \ref{fig:lastFixed}).
Figure \ref{fig:timeFirstFixed} and Figure \ref{fig:timeLastFixed} show the computation time graphs of the aforementioned setups with a maximum planning time of 1 second (for better visibility) and 30 seconds respectively.
We can see a significant difference in the planning duration until a path was found. 
The setup where the first contact point is in sliding contact was solved in under two seconds by all five algorithms.
In the other setup the path was found after a much longer time by three of the algorithms and the other two exceeded the maximum planning time.

This observed limitation may stem from the way the configurations are projected. In this work, we only consider setups in which the root of the robot is positioned on the robot's first contact joint. 
When we place the root on the robot's end joint that underlies a transition constraint, the configurations are projected differently, which affects the sampling of the transition constraint. Further research is needed to determine why a discrepancy between projections exist and which projection would be most useful for a specific robot and task.

Finally, we can also see that the required computation time differs depending on the scene setup.
We considered three kinds of objects in our scenarios: rectangles in the 2D scenarios, and cuboids and a sphere in the 3D scenarios.
The sphere in the 3D scene leads to a longer computation time than the cuboid and rectangular objects.
This can be attributed to a much higher number of surface polygons that has to be iterated over during the planning process. Note that analytical expressions exist for the sphere and using mesh polygons is not strictly necessary. However, our goal is to have a method which works for any mesh-based object representation. It is clear that a 3D scenario leads to more potential contact surfaces and thus also results in more polygons. To address this problem in future work, we believe it to be advantageous to use additional projection methods \cite{Orthey2020IJRR}, which could create equivalence classes of polygons.

%% file: sections/07conclusion.tex
\section{Conclusion}

In this work we contributed a geometric approach to planning sliding contacts using constrained planning and developed a sampler method for sliding transition constraints.
We formulated two constraints for a robot's contact joints: a sliding contact constraint and a transition constraint.
For the transition constraint we factored the problem into three constraint modes and implemented a sampling method that samples them uniformly.
We addressed the planning of a single contact joint breaking contact, moving it toward a goal and creating a new contact while the other joints remain in fixed or sliding contact.

Our tests and evaluations using sampling-based planning algorithms showed that the presented concept works on manipulator arms in 2D and 3D scenarios.
We observed that the planning time significantly depends on the projection used for the transition motion. To overcome this limitation, we believe it fruitful to further study the structure of the projection itself \cite{Berenson2011, Orthey2020IJRR} or to learn explicit representations of transition constraints \cite{Fernandez2020}.

Despite these limitations, we can successfully combine sliding contacts with transition constraints in a constraint-based sampling framework.
Using this approach we were able to solve planning problems for a robot with sliding contact points, which is fundamental and a first step towards probabilistically complete and optimal planning of long-horizon contact sequences which involve sliding contacts.